\documentclass{article}

\usepackage[backend=biber, style=numeric-comp, natbib=true]{biblatex}
\usepackage[sglblindworkshop, final, nonatbib]{neurips_2025}
\workshoptitle{Weather4Cast 2025}

\usepackage[utf8]{inputenc} % allow utf-8 input
\usepackage[T1]{fontenc}    % use 8-bit T1 fonts
\usepackage{url}            % simple URL typesetting
\usepackage{booktabs}       % professional-quality tables
\usepackage{amsfonts}       % blackboard math symbols
\usepackage{nicefrac}       % compact symbols for 1/2, etc.
\usepackage{microtype}      % microtypography
\usepackage{xcolor}         % colors
\usepackage{graphicx}

\usepackage{dsfont}
\usepackage{amsmath}
\usepackage[colorlinks=true, allcolors=blue]{hyperref}       % hyperlinks

\title{A Space-Time Transformer for Precipitation Nowcasting}

\author{
  Levi Harris \\
  UNC Chapel Hill \\
  \texttt{levlevi@cs.unc.edu} \\
  \And
  Tianlong Chen \\
  UNC Chapel Hill \\
  \texttt{tianlong@cs.unc.edu}
  }
\begin{document}

\maketitle

% --- outline ---
% 1. motivate need for reliable, in-situ (i.e., real-time) precipiation forecasts
    % planet is warming; extreme precipitation is becoming more common
    % meteorologists hoping to issue warnings and watches can't wait for the top of the hour model runs to come through?
% 2. traditional NWP models are expensive; difficult to quickly adopt to new regions; are run at fixed-intervals
% 3. previous regional-scale dl works for precipiation forecasting
    % 3a. rely on UNet/3D-convnet designs (TODO: validate)
    % 3b. are not well-calibrated to handle extremes (biased towards high-frequency, low-impact events)
% 4. we present SaTformer (a transformer model build on space-time attention modules designed for satallite videos)
% 5. we validate our approach by predicting distributions of cummulative rainfall over 64x64 km regions derived from OPERA radar rain-rate fields.
% 6. our method achieved 1st place at the NeurIPS Weather4Cast challenge 2025.

\begin{abstract}

% While AI-WP methods have enjoyed enormous success for medium range forcasting, applications of deep learning for nowcasting at fine-spatial resolutions are underexplored. 
    % rewrite this sentance; this is the core motivation for our paper

Until recently, numerical weather prediction (NWP) models have stood rivalless in operational forecasting despite a few limitations. Namely, physically-based models are computationally demanding and struggle at short lead times, reducing their utility for nowcasting. Motivated by these shortcomings, recent work proposes AI-weather prediction (AI-WP) alternatives that emulate analysis data with neural networks. While these data-driven approaches have achieved high skill for medium-range forecasting–applications of AI-WP to precipitation and to nowcasting are less explored. To these ends, this paper discusses \textit{SaTformer}: a video transformer adapted for precipitation nowcasting. To ameliorate some problems related to what is essentially a fat-tailed regression task, we find it prudent to formulate nowcasting as a classification problem and employ a frequency-weighted loss. This straightforward approach scored first on the NeurIPS Weather4Cast 2025 ``Cumulative Rainfall'' challenge. Code and model weights are available: \texttt{\href{github.com/leharris3/w4c-25}{github.com/leharris3/satformer}}.

% Meteorological agencies around the world rely on real-time flood guidance to issue live-saving advisories and warnings. 

% The abstract paragraph should be indented \nicefrac{1}{2}~inch (3~picas) on
% both the left- and right-hand margins. Use 10~point type, with a vertical
% spacing (leading) of 11~points.  The word \textbf{Abstract} must be centered,
% bold, and in point size 12. Two line spaces precede the abstract. The abstract
% must be limited to one paragraph.

\end{abstract}
\section{Introduction}

Extreme precipitation creates risk for life, property, and commerce. Recent studies predict that these events are likely to increase in frequency and severity during the coming decades, spurred on by a warming atmosphere \cite{Madakumbura_2021, Kotz_2022}. Consequently, members of the public and decision-makers require access to timely and accurate forecasts of rainfall at convective-allowing scales. For decades these data came from numerical models. But in recent years, a growing tide of AI research has reached the Earth sciences, helping spawn new techniques for precipitation nowcasting in stark contrast to the status quo. 

Recent data-driven AI-weather prediction (AI-WP) approaches offer a compute-efficient alternative to NWP models by learning to emulate forecast analysis data. Works from Google \cite{Lam2023, Price2024}, NVIDIA \cite{pathak2022fourcastnetglobaldatadrivenhighresolution}, Huawei \cite{Bi_2023}, and others replace core components of the NWP pipeline with neural networks \cite{abdi2025hrrrcastdatadrivenemulatorregional, Flora2025, Das2024, Asperti2024, Reulen2024, Ha2024}. Generative models have manifold advantages over numerical models. Namely, AI-WP pipelines can practically run at sub-hour resolution and in large ensembles; therefore, it is hopeful that operational forecasters can eventually leverage these models as a source of real-time guidance. However, previous studies primarily explore AI-WP for medium-range forecasting on a global scale. By contrast, precipitation nowcasting at continental and regional scales remains a more nascent challenge. 

This paper describes a solution to the NeurIPS Weather4Cast 2025 ``Cumulative Rainfall'' challenge. Here, we repurpose a video transformer architecture to predict future rainfall intensities from HRIT satellite radiances. We also describe a few obstacles we encountered adapting a model designed for classification to an imbalanced regression problem (rainfall intensities are fat tailed). 

\newpage

Motivated by these challenges, we seek answers to the following research questions:

\begin{enumerate}
    \item What modifications are required to adapt video transformers for precipitation nowcasting?
    \item Which techniques support data-driven model skill (e.g., overall accuracy, calibration, etc.) when training on fat-tailed precipitation datasets?
\end{enumerate}

To investigate the points above, we propose \textit{SaTformer}: a video transformer adapted for precipitation nowcasting using low-resolution satellite imagery as input. We find that with minimal modifications, video transformers provide a strong baseline despite label imbalances and heavily skewed training data. Our results support the hypothesis that data-driven models may be better suited for convective-scale nowcasting–where numerical models can balloon in complexity.

\begin{figure}[t]
    \centering
    \includegraphics[width=1.0 \linewidth]{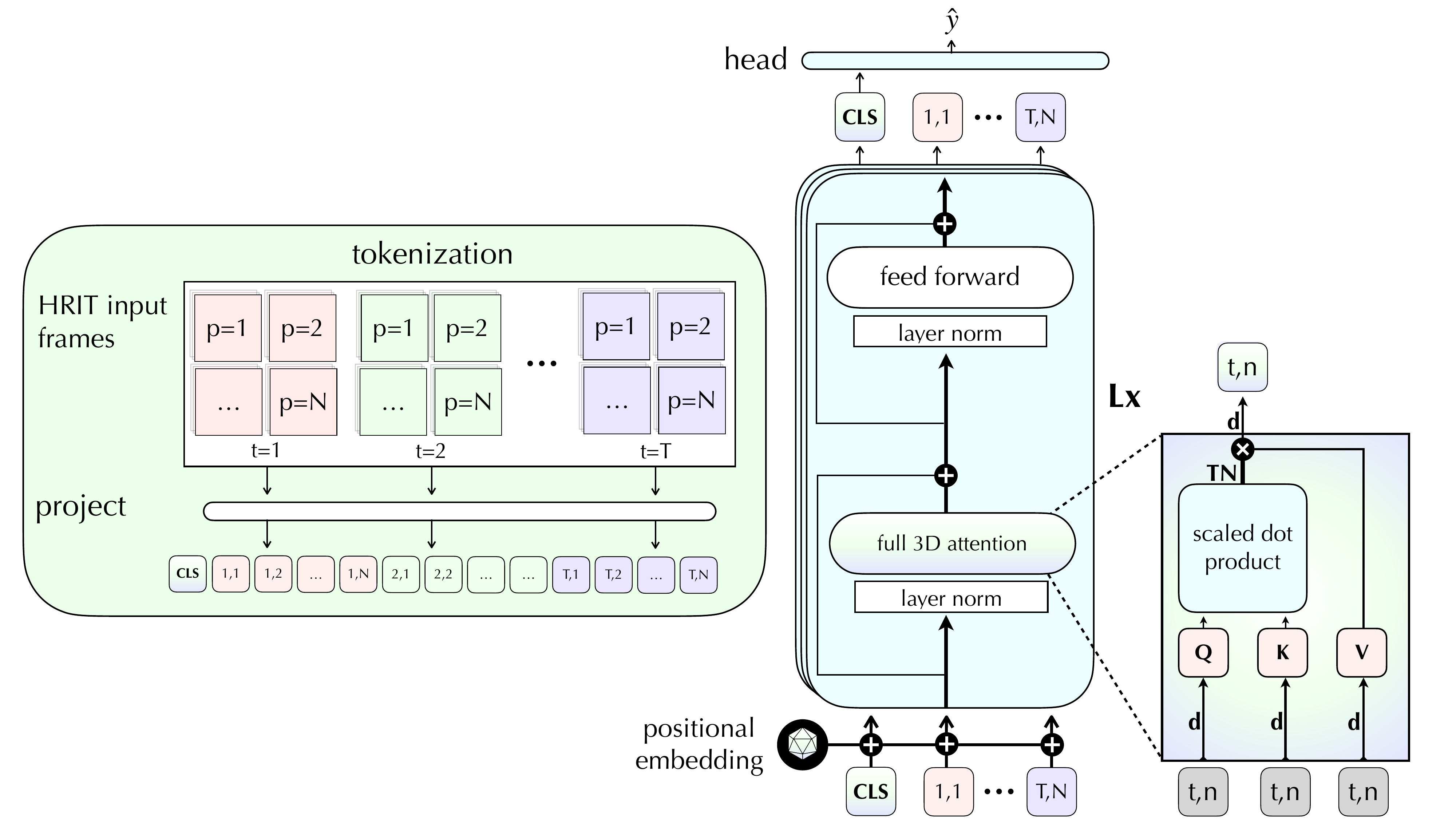}
    \caption{The SaTformer architecture. (Left) We partition each frame \(t \in (1, ...,T) \) in a sequence of satellite radiances into \( N \) non-overlapping patches. Each patch is projected into a token representation, and a class token (CLS) is prepended to the token sequence. (Center) Model encoder design. Token sequences pass through (L) transformer layers. The class token is spliced from the output of the final transformer layer and passed to a single-layer prediction head. (Right) Full 3D attention; all tokens attend to all tokens over time and space.}
    \label{fig:architecture}
    
\end{figure}

\section{Background}

\subsection{Weather4Cast competition}

Beginning in 2021, the annual Weather4Cast competition (W4CC) evaluates machine learning (ML) techniques for weather forecasting or precipitation nowcasting, with a particular focus on out of domain generalization \cite{pmlr-v220-gruca23a, gruca2024multitask}. Each year, participants are tasked with developing approaches to predict (OPERA) high-resolution radar precipitation fields (or targets derived from these data) using 11-band HRIT geostationary satellite data and static topographic maps as context. Competition data cover 10 non-overlapping regions (spatial areas of 1512\( \times \)1512 km) across Europe, northern Africa, and the Middle East for which HRIT data are completely available, while OPERA data are sparse.

\section{Related work}

Motivated by strong overlaps between subsequent task variations, we find it suitable to review works from all preceding years of the W4CC.

\subsection{W4CC 2021}

In its inaugural year, organizers proposed a multivariate weather forecasting challenge: participants predicted future atmospheric states (represented by four variables) for eight hours of lead-time using the same features and topographic information as model inputs \cite{kwok2021variationalunetweatherforecasting}. The authors of \cite{kwok2021variationalunetweatherforecasting} proposed to solve this task by training a U-Net VAE with a combination of deterministic and probabilistic losses. In this work, the authors argue that these design choices are crucial for modeling the inherent and model-related uncertainty present in weather forecasting (i.e., aleatoric and epistemic). However, this complex, multi-loss setup may prove difficult to extend to new tasks in practice.

\subsection{W4CC 2022}

In 2022 the multi-variate challenge from 2021 was replaced by a univariate, binary (rain/no-rain) precipitation nowcasting task defined over 64\(\times\)64 km OPERA radar-regions and 8 hours of lead-time.

In \cite{belousov2022solvingweather4castchallengevisual} an ensemble of video classification models (e.g., Swin UNETR, ViViT) is used \cite{hatamizadeh2022swinunetrswintransformers, arnab2021vivitvideovisiontransformer}. An obvious drawback is that this approach requires training three different models from scratch for a single task and incurs further computational costs by bootstrapping predictions at inference-time. Additionally, modifications made by the authors to adapt video classification models to a video generation task (e.g., appending a single convolutional layer to ViViT to predict the output temporal dimension) are likely brittle design choices. In \cite{li2022superresolutionprobabilisticrainprediction}, the authors apply a 3D U-Net and an Earthformer \cite{çiçek20163dunetlearningdense, gao2023earthformerexploringspacetimetransformers} to solve this year's stage 1 and stage 2 challenges, respectively. In the second challenge, an Earthformer is employed to efficiently decompose a video-sequence into cuboids (as opposed to frame-wise patches as proposed in ViT); self-attention is performed between cuboids and a collection of global context tokens \( \mathcal{G}\). However, inter-cuboid attention is axial, and therefore, cuboids only learn spatial dependencies from a set of coarse global context vectors.

Beginning with a standard 3D U-Net, the authors of \cite{park2022rainunetsuperresolutionrainmovie} implement a time-space (TS) convolutional block in which a spatial convolution, followed by a ``temporal-wise dilated convolution'' (TSDConv), and finally a temporal convolution are applied to the input feature maps. The authors note that their proposed TSDConv is simply an extension of the large kernel convolution operation into three dimensions. Additionally, WeatherFusionNet \cite{pihrt2022weatherfusionnetpredictingprecipitationsatellite} proposes a physically informed U-Net-based architecture. As inputs, their model uses HRIT satellite radiances, a PDE-constrained video prediction via PhyDNet \cite{guen2020disentanglingphysicaldynamicsunknown}, and the outputs of a proposed Sat2Rad module to forecast binary rain masks. While the methods above achieve strong performance on this year's tasks, most are fundamentally reliant on CNN model designs, and likely suffer from the inductive biases and poor scaling these architectures exhibit in other domains.

\subsection{W4CC 2023}

Tasks did not change between 2022 and 2023.

This year, researchers at Alibaba Cloud proposed a mixture-of-experts (MoE) inspired solution to both the ``core'' and ``transfer learning'' competition tracks \cite{li2023precipitationpredictionusingensemble}. The authors design a ``predictor'', which samples multiple times from a WeatherFusionNet backbone, and a ``controller''; the latter module produces a final output as a weighted combination of the expert predictions. In addition to being an extension of prior work, the proposed model employs a slightly cumbersome three-stage training scheme. Their approach is still preferable to \cite{kumar2023skilfulprecipitationnowcastingusin}, who simply finetune a NowcastNet \cite{Zhang2023} on the challenge dataset. In \cite{han2023learningrobustprecipitationforecaster} an augmentation technique is proposed to sample linear interpolations of the original training samples. By formulating precipitation forecasting as a multi-class, video classification task (where each per-pixel class represents a discrete interval of precipitation), the authors are able to employ a modified dice-loss with an added ordinal prior (i.e., under-predictions are punished more heavily than over-predictions). While certainly performant, this approach is heavily engineered and therefore may generalize poorly to new tasks and datasets.

\subsection{W4CC 2024}

In 2024, the W4CC organizers shortened the duration of target OPERA sequences from 8 hours to 4 hours.

In \cite{deshpande2024conditionalgenerativeadversarialnetwork}, the authors develop a conditional GAN network trained on a sparsely-sampled subset of the original HRIT channels. Here, the authors discard near-infrared, water vapor, and visible satellite channels, and average the remaining 4 channels; after data preprocessing their model is trained on only single-channel inputs. Further preprocessing techniques are employed to improve model performance, including thresholding out non-cloudy regions and extracting optical flow scores from the input sequence. Similar to \cite{han2023learningrobustprecipitationforecaster}, this work proposes a bespoke, highly-engineered pipeline that is unlikely to extend gracefully to future work.
% >*A space-time transformer for precipitation regression*

% Pipeline
% - **Figure**: show input tokenization process
% 	- A multichannel satellite radiance (T, C=11, H, W) is patchified -> (T, N, 11 * P^2)
% 	- Satellite patches are projected down into tokens -> (T, N, D)
% - In practice: we combine the T and N dims -> (T * N, D)
% - Apply positional embeds
% 	- *Note*: we use **rotary embeds**
% - Prepend a `cls` token -> ((T * N) + 1), D)
% - Iteratively apply mutli-head space -> time attention
% 	- **Figure**: visualize space-time decoupled attention
% - Use a (D, N) linear layer to predict output classes from `D` shaped `cls` token

% Misc
% - **Figure**: visualize a target OPERA radar sequence
% 	- *Include derivation of the ground truth regression targets*
% - **Figure**: histogram of the target distribution with a **log scale** on the y axis
% - **Figure**: ablation
% 	- `CRPS`?
% 	- 1. model performance w/ regression only (i.e., 1 output neuron)
% 	- 2. model performance w/ categorical (CCE)
% 	- 3. model performance w/ class-weighted cce
% 	- **NOTE**: we need to choose cls-balanced/extremes-weighted metrics (a model that achieves good performance by over-predicting low-precip isn't useful in our case)

% Ideas
% - If there's time, play around with full O(N^3) attention*

% A. Model: SaTformer
    % 1. input patchification
    % 2. positional (rotary)* embeddings
    % 3. transformer module
        % 3a. full space-time attention 
    % 4. classification layer
% B. Regression -> categorical task formulation
% C. Loss: weighted-categorical cross entropy

\section{Method}

\subsection{SaTformer model}

% \to \mathbf{y} \)

Taking inspiration from video understanding literature, we implement a transformer-based architecture for precipitation nowcasting (Fig. \ref{fig:architecture}). Given one hour of low-resolution HRIT satellite data \( \mathbf{x} \in \mathbb{R}^{T \times C \times H \times W} \), we train a model to learn a mapping to a cumulative distribution of rainfall targets at a single point \(F(y_{\text{reg}}) \), where \(y_{\text{reg}}\) is the target label provided by the organizers.

Following prior work \cite{dosovitskiy2021imageworth16x16words, bertasius2021spacetimeattentionneedvideo}, we partition each frame (\(T\)) in the input sequence into \( N \) non-overlapping patches with height \(P\): \( \mathbf{x}^{(0)}_{(t, p)} \in \mathbb{R}^{P \times P \times C} \); patches completely cover the spatial dimensions of the original frame. Next, we flatten each patch and project the resulting vectors into tokens: \( \mathbf{z'}^{(0)}_{(t, p)} \in \mathbb{R}^{d} \). We prepend a randomly initialized ``class token'' \(\mathbf{z'}_{(0, 0)}^{(0)}\) to the token sequence intended to aggregate relevant information for prediction of the target label:

\begin{equation}
    % [\mathbf{z'}_{(0, 0)}^{(0)}, \mathbf{z'}^{(0)}_{(0, 1)}, \dots, \mathbf{z'}^{(0)}_{(0, N)}, \mathbf{z'}^{(0)}_{(1, 0)}, \dots, \mathbf{z'}^{(0)}_{(T, N)}]
    [\underbrace{\mathbf{z'}^{(0)}_{(0,0)}}_{\text{cls token}}, \mathbf{z'}^{(0)}_{(0, 1)},\dots, \mathbf{z'}^{(0)}_{(T, N)}].
\end{equation}

We then add learnable positional embeddings \( \mathbf{e}_{(t, p)} \in \mathbb{R}^d \) to all \(NT+1\) tokens in the input sequence:

% \begin{align}
%         \mathbf{z}^{(0)}_{(0, 0)} = \mathbf{z'}^{(0)}_{(0, 0)} + 
%         \mathbf{e}_{(0, 0)} &&
%         \mathbf{z}^{(0)}_{(t, p)} = \mathbf{z'}^{(0)}_{(t, p)} + \mathbf{e}_{(t, p)}.
% \end{align}

\begin{equation}
\begin{alignedat}{2}
\mathbf{z}^{(0)}_{(0,0)}
&= \mathbf{z'}^{(0)}_{(0,0)} + \mathbf{e}_{(0,0)},
\qquad&
\mathbf{z}^{(0)}_{(t,p)}
&= \mathbf{z'}^{(0)}_{(t,p)} + \mathbf{e}_{(t,p)}.
\end{alignedat}
\tag{2,3}
\end{equation}

\subsubsection{Space-time self-attention}

Our model is composed of \(L\) transformer encoder blocks that sequentially operate on the input token sequence. To capture rich, long-distance dependencies over time and space, we implement full space-time self-attention. Here, each token in the input sequence is mapped to a query, key, and value vector using respective weight matrices \(W \in \mathbb{R}^{d \times d}\):

\setcounter{equation}{3}
\begin{align}
    \mathbf{q}^{(l)}_{(t, p)} &= 
    \text{LN}(\mathbf{z}^{(l-1)}_{(t, p)}) W_{q}, &
    \mathbf{k}^{(l)}_{(t, p)} &= \text{LN}(\mathbf{z}^{(l-1)}_{(t, p)}) W_{k}, &
    \mathbf{v}^{(l)}_{(t, p)} &= \text{LN}(\mathbf{z}^{(l-1)}_{(t, p)}) W_{v},
    \tag{4,5,6}
\end{align}

\setcounter{equation}{6}

where \( \mathbf{q}^{(l)}_{(t, p)} \in \mathbb{R}^{d} \),  \( \mathbf{k}^{(l)}_{(t, p)} \in \mathbb{R}^{d} \), \( \mathbf{v}^{(l)}_{(t, p)} \in \mathbb{R}^{d} \) and \(\text{LN}(\cdot)\) denotes the LayerNorm operation \cite{ba2016layernormalization}. Next we compute importance weights \( \mathbf{a}^{(l)}_{(t, p)} \) between all pairs of tokens across time and space:

\begin{equation}
    % \mathbf{a}^{(l)}_{(t, p)} = \text{softmax} \big{(} \frac{(\mathbf{q}^{(l)}_{(t, p)})^{\top}}{\sqrt{d}} \cdot [\mathbf{k}_{(0, 0)}^{(l)}, \mathbf{k}^{(l)}_{(0, 1)}, \dots, \mathbf{k}^{(l)}_{(0, N)}, \mathbf{k}^{(l)}_{(1, 0)}, \dots, \mathbf{k}^{(l)}_{(T, N)}]\big{)}
    \mathbf{a}^{(l)}_{(t, p)} = \text{softmax} \big{(} \frac{(\mathbf{q}^{(l)}_{(t, p)})^{\top}}{\sqrt{d}} \cdot [\mathbf{k}_{(0, 0)}^{(l)}, \mathbf{k}^{(l)}_{(0, 1)}, \dots, \mathbf{k}^{(l)}_{(T, N)}]\big{)},
\end{equation}

where \( \mathbf{a}^{(l)}_{(t, p)} \in \mathbb{R}^{TN+1} \). The output of the self-attention layer is a weighted sum calculated between all value vectors and importance weights for each token:

\begin{equation}
    \mathbf{s}^{(l)}_{(t, p)} = \mathbf{a}^{(l)}_{(t, p)(0, 0)} \mathbf{v}^{(l)}_{(0, 0)} + \sum_{t' = 1}^{T} \sum_{p' = 1}^{N}  \mathbf{a}^{(l)}_{(t, p)(t', p')} \mathbf{v}^{(l)}_{(t', p')},
\end{equation}

where \( \mathbf{s}^{(l)}_{(t, p)} \in \mathbb{R}^{d} \). The resulting attention scores are projected and summed with a residual connection from the previous encoder block; these resulting vectors are passed through the remainder of the transformer layer:

\begin{align}
    \mathbf{z'}^{(l)}_{(t, p)} = 
    \mathbf{s}^{(l)}_{(t, p)} W_\text{out} + \mathbf{z}^{(l-1)}_{(t, p)},
    &&
    \mathbf{z}^{(l)}_{(t, p)} = \text{MLP}(\text{LN}( \mathbf{z'}^{(l)}_{(t, p)})) + \mathbf{z'}^{(l)}_{(t, p)}.
    \tag{9, 10}
\end{align}

\setcounter{equation}{10}

Note that in practice we employ a standard transformer encoder-design implemented with multi-headed attention; notation for multi-headed attention is omitted for brevity.

\subsubsection{Categorical task formulation}

Deep neural networks often struggle to match the performance of ML baselines on regression tasks, particularly for imbalanced data regimes \cite{grinsztajn2022treebasedmodelsoutperformdeep, yang2021delvingdeepimbalancedregression}. To address these challenges and to exploit proven model designs for classification tasks, we reformulate precipitation regression as a classification problem. Given a target regression label \(y_{\text{reg}}\) provided by the organizers and a training dataset \(\mathcal{D}\), we approximate categorical equivalent labels \(y\) by partitioning the target space into \( n \) non-overlapping bins spaced by step size \( \delta \):

\[
(\mathcal{D}_{\text{ymin}}, \mathcal{D}_{\text{ymin}} + \delta, ...,\mathcal{D}_{\text{ymin}} + i\delta , ..., \mathcal{D}_{\text{ymax}})_{i=0...n-1}.
\]

We then generate one-hot categorical labels \( y \in \{ 0, 1\}^n\), representing a bin centered at \( \mathcal{D}_{\text{ymin}} + i\delta\), by calculating the nearest label-bin to a target: \( i = \text{round}(\frac{y_{\text{reg}} - \mathcal{D}_{\text{ymin}}}{\delta} )\). Our model learns to predict categorical labels derived above. Taking the output of the last transformer encoder block, we splice out the last class token \( \mathbf{z}^{(L)}_{(0, 0)}\); a one-layer MLP prediction head generates class-probabilities using this token as input:

\begin{equation}
    \hat{y} = \text{softmax} \big{(} \text{MLP}( \mathbf{z}^{(L)}_{(0, 0)}) \big{)}.
\end{equation}

\subsubsection{Class-weighted cross-entropy loss}

Extreme precipitation events are rare events by nature. Standard classification loss functions (e.g., cross-entropy) bias models in favor of the majority class, degrading performance on long-tailed datasets like ours (Fig. \ref{fig:label-distribution}). We employ a reweighted categorical cross entropy loss to offset class imbalances and promote model skill when forecasting outlier events:

\begin{equation}
    \label{w-cce}
    % \mathcal{L}(y,\hat{y}) = -\sum_{i=0}^{n-1} w_i \log(\hat{y}_i)y_i
    \mathcal{L}(y,\hat{y}) = -\sum_{i} w_i \log(\hat{y}_i)y_i.
\end{equation}

% note: our loss fn weighting term w is actually just the self-information I(X=cls-n) = -log(p(X=cls-n))
% https://www.google.com/url?sa=t&source=web&rct=j&opi=89978449&url=https://en.wikipedia.org/wiki/Information_content&ved=2ahUKEwjsoteUo4KSAxXYl2oFHclqL8MQFnoECB8QAQ&usg=AOvVaw2NzFzHTCXw-HwIHrOmvxNw

Here \( \hat{y} \) is our model's predicted probability distribution over the target classes and \(w_i\) is the inverse frequency of a class in the training set.

%  = -\log(\frac{|\mathcal{D}_i|}{|\mathcal{D}_{\text{total}}|})\)

% TODO: resolve some ambiguity here
% TODO:
% * Here we use 'N' to denote patches/frame; make sure we don't abuse

% (\textbf{TODO: FIX THIS}) \( W_{\text{emb}} \in \mathbb{R}^{d \times N} \):

% MH-full attention -> BN -> MLP -> BN
% * attention for cls token computed separately...relevant for us?
% * TODO: we use Lx at different points in the paper to denote model depth; switch -> L everywhere

% TODO:
% * currently z^(l) is the input token to an encoder; not the OUTPUT token
% TODO: verify this is equivalent to a @ v

% \bigg{[}[i *  \delta, (i+1) * \delta) \bigg{]}_{i=0...n-2}
% [0, i\delta ), [\delta, 2\delta), ..., [(n-1)\delta, n\delta)
% [y_{\text{cat}_1}, y_{\text{cat}_2}), [y_{\text{cat}_2}, y_{\text{cat}_3}), ..., [y_{\text{cat}_{n-1}}, y_{\text{cat}_n}]
\section{Experiments}

% TODO: ABLATIONS

% 0. begin with an unmodified TimeSformer

% 1. bin-averaged mROI
    % w/  class-weighted loss
    % w/o class-weighted loss
% 2. attention
    % 1. full space time  attention
    % 2. space   -> time  attention
    % 3. time    -> space attention
% 3. num_output_classes
    % since we optimize to the test set, include these ablations last

% * metrics
    % bin-averaged CSI
    % bin-averaged F1
    % bin-averaged CRPS

% 1. dataset pre-processing
    % a. input/output normalization
    % b. sampling
        % b1. select a random region; select a random timestep; select 4 frames of HRIT data as input; select 16 frames of OPERA data as output
        % b2. select a random 32x32 pixel (i.e., 392x392 km crop) from HRIT
        % b3. select a 32x32 pixel (i.e., 64x64 km) crop of OPERA data with the same spatial center as the HRIT data
            % *we might condider showing how this is done in the appendix
        % b4. generating target labels
            % refer back to earlier derivation of targets
% 2. training details
    % batch size: 128 effective
    % GPUs: 1xA6000
    % optimizer: Adam
    % LR: 1e-3
    % epochs: 20 or until convergence
    % steps per epoch: 8191
% 3. ablations study
    % 3a. [VAL ] loss formulation
        % bin-weighted (bw) F1 (bw-F1); bw-
    % 3b. [VAL ] attention variants
    % 3c. [TEST] num_output_classes
        % [FIGURE]: line plot
        % since we optimize to the test set, include these ablations last
% 4. results
    % our method achieved first place at the WFCC 2025 with a score of 3.1

% misc
    % [FIGURE]: bar chart - emperical distribution of test-set target values

\subsection{W4CC 2025: cumulative rainfall estimation}

We apply our model to the ``Cumulative Rainfall'' task from W4CC 2025. This year's task involved prediction of cumulative precipitation targets from 11-band satellite imagery spanning visible, infrared, and water vapor channels. Given 1 hour of HRIT satellite imagery, \(\mathbf{x}\), target labels \( y_{\text{reg}} \) are derived from a 4 hour sequence of OPERA high-resolution rain rate fields sharing the same spatial center point. The organizers state this target can be derived by: ``[averaging] over a 32×32 pixel area...averaging the 16 slots and multiplying by 4 (or summing the 16 slots and diving [sic] by 4)''.

% \begin{equation}
%     y_{\text{reg}} = \frac{4}{T'W'H'}\sum_{t=1}^{T'}\sum_{i=1}^{H'}\sum_{j=1}^{W'} \mathbf{r}_{(t, i, j)}
%     \label{fig:target-derive}
% \end{equation}

\subsection{Dataset}

We train our model using 7 region-specific datasets provided by the W4CC organizers containing pairs of HRIT satellite radiances and OPERA precipitation fields. At each step in our training process, we select a random region from our dataset. Within this region, we randomly sample a 32\(\times\)32 pixel (512\(\times\)512 km) spatial crop of HRIT data for a random, four-consecutive frame (1 hour) temporal window. This subsample comprises our model input. We then select a 32\(\times\)32 pixel (64\(\times\)64 km) OPERA sample centered at the same geographic point and spanning the 16 frames (4 hours) following the model input data. From this radar field we derive categorical target labels. Next, we normalize all input features roughly between \([0, 1]\) using statistics calculated over our training set.

% (Readers are referred to the appendix for full details of our sampling strategy, including calculating projections from HRIT to OPERA space).

% \begin{align*}
%         \mathbf{x} = \frac{\mathbf{x}_{\text{raw}} - \mathcal{D_{\text{xmin}}}}{\mathcal{D_{\text{xmax}}} - \mathcal{D_{\text{xmin}}}}
%         % && y = \frac{y_{\text{raw}} - \mathcal{D_{\text{ymin}}}}{\mathcal{D_{\text{ymax}}} - \mathcal{D_{\text{ymin}}}}
% \end{align*}

\subsection{Training details}

We configure our SaTformer with input height and width 32, channel dimension 11, number of classes 64, patch size 4, hidden dimension 512, head dimension (for multi-headed attention) 64 using 8 heads, total transformer encoder blocks (L=12), initialized with learnable positional embeddings. We train our model for 200 epochs and a total of 25,000 steps using the Adam optimizer configured with a learning rate 1e-5, and an effective batch size 128 on 1\(\times\)A6000 GPU. We evaluate model checkpoints that score the lowest average validation loss during a training run.

% dumb and wrong: Our choice of patch size is inspired by ViT, who use a \(P\) value 14 for image side-lengths 224; our ratios of input side-length to \(P\) are therefore identical theirs at 1:8.

\subsection{Results}

All models entered in W4CC 2025 are evaluated using continuous ranked probability score (CRPS). For each input sequence in the test set, the model-predicted cumulative distribution function \(F\) is compared against the ground-truth observation \(y_{\text{reg}}\):

\begin{equation*}
\text{CRPS}(F, y_{\text{reg}}) = \int_{- \infty}^{\infty} \big{(} F(x) - \mathds{1} \{x - y_{\text{reg}}\}\big{)}^2 dx.
\end{equation*}

% \begin{equation*}
% \text{CRPS}(F, y) = \int_{- \infty}^{\infty} \big{(} F(y) - \mathds{1} (\hat{y} - y)\big{)}^2 d\hat{y}.
% \end{equation*}

Our best performing model achieves a CRPS score of 3.135 on the challenge set, scoring first in the W4CC 2025 ``Cumulative Rainfall'' challenge.

\subsection{Ablation study}
% NOTE: an ablation validates that REMOVING a component/model feature DEGRADES performance

\begin{figure}[t]
    \centering
    \includegraphics[width=1.0 \linewidth]{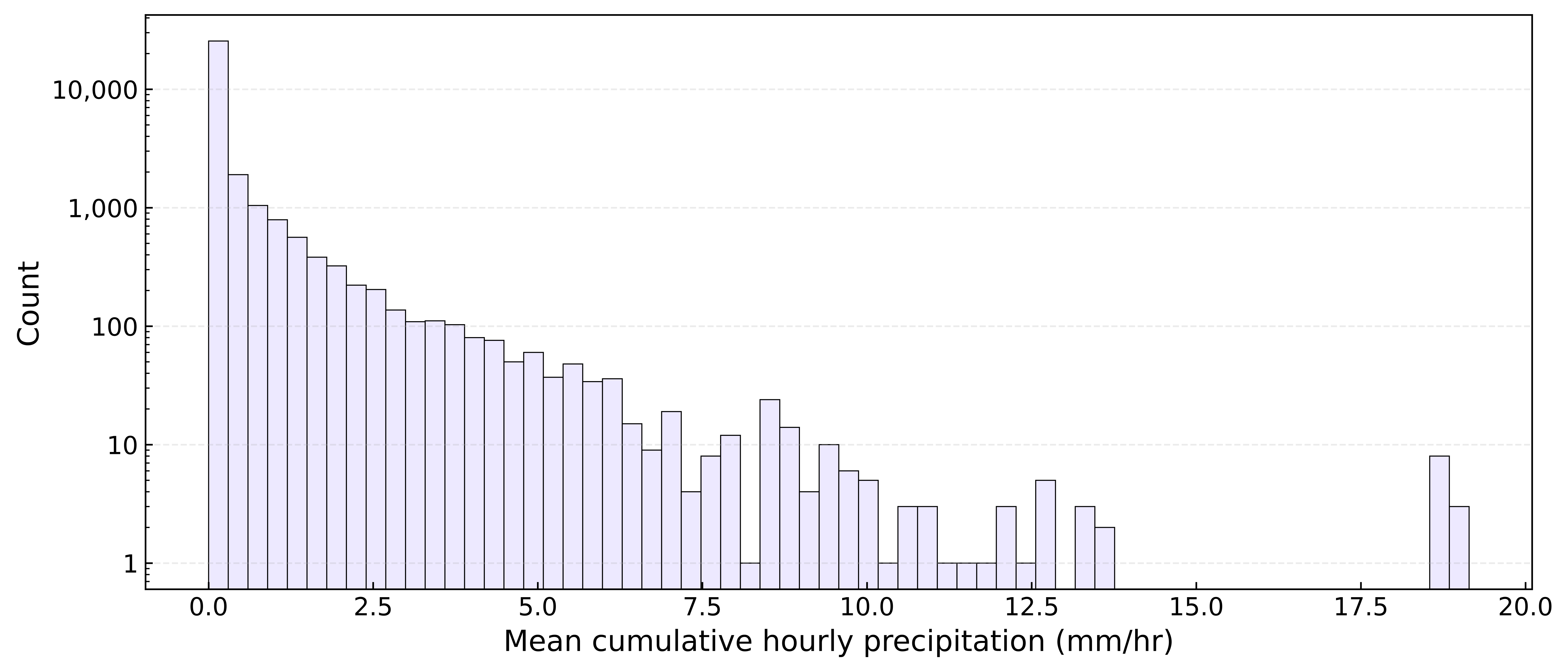}
    \caption{Empirical distribution of target labels within our training set. Note the log scale on the y-axis; targets for our task are skewed heavily towards low/no rain events.}
    \label{fig:label-distribution}
\end{figure}

\begin{table}[t]

    \centering
    
    \caption{Ablations conducted on the W4CC validation set. To assess model skill on poorly represented classes (i.e., extreme precipitation events) we develop bin-weighted (BW) variants of standard regression and classification metrics. Here, we report class averaged model accuracy (BW-Top-3) and CRPS scores. (Top) We compare model performance with and without class-frequency re-weighting. When using a standard CCE loss, our model quickly overfits to low/no-rainfall events. (Bottom) Comparison of three attention variants: space followed by time (S\(\to\)T), time then space (T\(\to\)S), and full space-time self-attention (S + T).}
    
    \begin{tabular}{c c c c}
    \toprule
    
    \textbf{Loss weighting?}            & 
    \textbf{BW-Top-3} \(\uparrow \)     &
    \textbf{BW-CRPS}  \(\downarrow \) \\
    
    \midrule
    -          & 0.076        & 6.91 \\
    \checkmark & \textbf{0.272}        & \textbf{2.64} \\
    
    \midrule

    \textbf{Attention}        &
                              &
                              \\
    \midrule

    S\(\to\)T & 0.250 & 4.39 \\
    T\(\to\)S & 0.214 & 3.39 \\
    S + T     & \textbf{0.272} & \textbf{2.64} \\

    \bottomrule
    
    \end{tabular}
    
    \label{tab:loss-and-attention-ablation}
    
\end{table}

We conduct ablations for several design choices made throughout this paper. Concretely, our SaTformer can be derived step-by-step from a TimeSformer baseline \cite{bertasius2021spacetimeattentionneedvideo} implemented with interleaved space-time attention modules and trained via standard CCE loss. As previously mentioned, datasets for precipitation nowcasting commonly suffer from large class imbalances. As shown in Figure \ref{fig:label-distribution}, our target distribution is heavily skewed towards low/no rain events and contains large gaps in sample coverage at the extremes. To encourage balanced performance across classes we implement the weighted CCE loss proposed in equation \eqref{w-cce}; ablations performed using bin-averaged metrics confirm that including a class-frequency prior from the training distribution aids model skill (Table \ref{tab:loss-and-attention-ablation}).

We depart from the original TimeSformer paper, however, by implementing full attention in our transformer encoder layers. Complex video reasoning tasks (e.g., action recognition) typically require high-dimensional inputs. However, the input satellite radiances provided to us by the organizers span just 4 frames, and require attention over a much smaller pixel-area than is typical for natural videos. Full space-time attention may allow our model to capture rich, long-distance dependencies over space and time. Ablations comparing various attention implementations confirm that full space and time attention yields the best model performance (Table \ref{tab:loss-and-attention-ablation}) on our cumulative precipitation estimation task.

Satisfied with our choice of loss function and encoder design, we conclude with evaluations to determine optimal target-label bin-size using the W4CC challenge set. We note that while choosing to predict more output classes (i.e., using smaller values of \(\delta\)) generally yields better CRPS scores in theory, in practice models tend to produce degenerate solutions as class representation becomes increasingly sparse. A reasonable middle ground of 64 classes achieves the best empirical results (Table \ref{tab:nbins-ablation}).

\begin{table}[h]
    \centering
    \begin{tabular}{c c}
    \toprule
    
    \textbf{\# Bins}                  & 
    \textbf{CRPS}  \(\downarrow \)      \\
    
    \midrule
    4  &                     14.181 \\
    8  &                     5.987  \\
    16 &                     4.293  \\
    32 &                     3.898  \\
    \textbf{64} &   \textbf{3.135}  \\
    128 &                    3.610  \\
    256 &                    5.312  \\
    512 &                    4.783  \\
    
    \bottomrule
    \end{tabular}
    \\[10pt]
    \caption{W4CC challenge set scores for increasingly fine-grained precipitation buckets.} 
    \label{tab:nbins-ablation}
\end{table}
\section{Discussion}

\subsection{RQs}

We briefly return to the research questions posed at the beginning of this paper. 

\textbf{``What modifications are required to adapt video transformers for precipitation nowcasting?''}

We find that video transformer models perform surprisingly well on a precipitation nowcasting task with minimal modification. Empirically, we find that we can improve upon model baselines by ablating the number of target label bins (Table \ref{tab:nbins-ablation}). Another task-specific modification we make is to implement full self-attention over space and time. This computationally expensive design-choice is made possible due to the small spatial dimensions of our input sequences. Researchers hoping to apply transformers to higher dimensional weather data will almost certainly require alternative methods that scale sub-quadratically with respect to the size of the input sequence.

\textbf{``Which techniques support data-driven model skill (e.g., overall accuracy, calibration, etc.) when training on fat-tailed precipitation datasets?''}

As shown in Table \ref{tab:loss-and-attention-ablation}, partitioning the target space into discrete bins allows us to achieve decent performance when applying a video classification model to precipitation nowcasting. Moreover, we observe that without additional calibration or class-regularization techniques, model performance can degrade towards degenerate solutions (i.e., always predict `no rain') in imbalanced learning regimes. Our experiments showed that a simple class-weighted cross entropy loss can achieve meaningful gains compared to an unweighted CCE. Taming fat-tailed datasets like those explored in this paper remains a critical challenge in AI-WP. Future work may improve upon the results presented here by employing data augmentation techniques or by utilizing other task formulations.

\subsection{Limitations}

\textit{SaTformer} achieves decent performance on a precipitation nowcasting task. However, significant modifications to this architecture will be required to produce dense nowcasts (i.e., video-like outputs) over space and time. Researchers may consider using graph networks or diffusion-based approaches that are better suited to generative prediction instead. Additionally, full self-attention (while a powerful mechanism for capturing space-time dependencies) incurs quadratic costs with respect to the length of the input sequence. Those interested in adapting our method to longer satellite videos may consider other efficient self-attention approaches (e.g., decoupled space-time attention, sparse self-attention, etc.). Future work may also benefit by employing a self-supervised encoder to compress video inputs prior to the tokenization stage. 

\section{Conclusion}

In this paper, we discussed SaTformer: an application of a video transformer that can skillfully predict future precipitation intensities using low-resolution satellite imagery as input. We found that by combining a categorical task formulation and a class-weighted loss function, we can offset some of the challenges encountered in an imbalanced training paradigm. Moreover (and somewhat unsurprisingly), full space-time attention improved model performance on our task, and thus may be a viable option for others so long as their training data is relatively low-dimensional.

\newpage

%%%%%%%%%%%%%%%%%%%%%%%%%%%%%%%%%%%%%%%%%%%%%%%%%%%%%%%%%%%%

% \input{docs/8_checklist}

\printbibliography

% \newpage
% \input{docs/7_appendix}

\end{document}